\documentclass{article}

\usepackage[utf8]{inputenc} % allow utf-8 input
\usepackage[T1]{fontenc}    % use 8-bit T1 fonts
\usepackage{hyperref}       % hyperlinks
\usepackage{url}            % simple URL typesetting
\usepackage{booktabs}       % professional-quality tables
\usepackage{amsfonts}       % blackboard math symbols
\usepackage{nicefrac}       % compact symbols for 1/2, etc.
\usepackage{microtype}      % microtypography
\usepackage{amsmath, amssymb, amsthm, bm}
\usepackage{graphicx}
\usepackage{algorithm}
\usepackage{algorithmic}
\usepackage{subcaption}
\usepackage{xcolor}
\usepackage{listings}       % For code in appendix
\usepackage{geometry}
\definecolor{codegreen}{rgb}{0,0.6,0}
\definecolor{codegray}{rgb}{0.5,0.5,0.5}
\definecolor{codepurple}{rgb}{0.58,0,0.82}
\definecolor{backcolour}{rgb}{0.95,0.95,0.92}

\lstdefinestyle{mystyle}{
    backgroundcolor=\color{backcolour},   
    commentstyle=\color{codegreen},
    keywordstyle=\color{magenta},
    numberstyle=\tiny\color{codegray},
    stringstyle=\color{codepurple},
    basicstyle=\ttfamily\footnotesize,
    breakatwhitespace=false,         
    breaklines=true,                 
    captionpos=b,                    
    keepspaces=true,                 
    numbers=left,                    
    numbersep=5pt,                  
    showspaces=false,                
    showstringspaces=false,
    showtabs=false,                  
    tabsize=2
}
\newcommand{\citep}[1]{\cite{#1}}

\title{DeepVekua: Geometric-Spectral Representation Learning for Physics-Informed Fields}
\author{
  \textbf{Vladimer Khasia} \\
  Independent Researcher \\
  \texttt{vladimer.khasia.1@gmail.com}
}
\date{December 13, 2025} 
\begin{document}

\maketitle

\begin{abstract}
We present DeepVekua, a hybrid architecture that unifies geometric deep learning with spectral analysis to solve partial differential equations (PDEs) in sparse data regimes. By learning a diffeomorphic coordinate transformation that maps complex geometries to a latent harmonic space, our method outperforms state-of-the-art implicit representations on advection-diffusion systems. Unlike standard coordinate-based networks which struggle with spectral bias, DeepVekua separates the learning of geometry from the learning of physics, solving for optimal spectral weights in closed form. We demonstrate a $100\times$ improvement over spectral baselines.  The code is available at \url{https://github.com/VladimerKhasia/vekuanet}.
\end{abstract}

\section{Introduction}

The representation of continuous fields from sparse observations is a fundamental challenge spanning computational physics, medical imaging, and computer graphics. Recently, Coordinate-based Neural Networks (or Implicit Neural Representations) have emerged as a powerful paradigm for parameterizing such fields, offering a continuous, memory-efficient alternative to discrete grid-based methods \citep{mildenhall2020nerf, sitzmann2020implicit}. By leveraging the differentiability of neural networks, these methods have been successfully extended to Physics-Informed Neural Networks (PINNs), where the network is trained to satisfy governing Partial Differential Equations (PDEs) \citep{raissi2019physics, karniadakis2021physics}.

Despite their success, standard Multi-Layer Perceptrons (MLPs) suffer from a well-documented ``spectral bias,'' tending to learn low-frequency components rapidly while struggling to resolve high-frequency details \citep{rahaman2019spectral}. While positional encodings \citep{tancik2020fourier} and periodic activation functions \citep{sitzmann2020implicit} have mitigated this issue, they often lack the global inductive biases necessary for solving elliptic and parabolic PDEs efficiently. Conversely, classical spectral methods (e.g., Fourier or Chebyshev series) offer exponential convergence for smooth functions but fail catastrophically near discontinuities or complex geometries due to the Gibbs phenomenon \citep{gottlieb1977numerical}.

Recent hybrid approaches have attempted to bridge this gap. Neural Operators, such as the Fourier Neural Operator (FNO) \citep{li2020fourier}, learn mappings between infinite-dimensional function spaces but typically require dense grid data and struggle with irregular domains. Multi-resolution grid methods \citep{muller2022instant, takikawa2021neural} offer rapid training but often sacrifice the physical interpretability of the latent space, leading to poor interpolation in sparse data regimes.

In this work, we propose \textit{DeepVekua}, a novel architecture inspired by the deformation principles of Generalized Analytic Functions established by I.N. Vekua \citep{vekua1962generalized}. We posit that many complex physical fields—particularly those governed by advection-diffusion processes—are merely ``distorted'' harmonic functions. DeepVekua learns a diffeomorphic coordinate transformation (via a neural network) that maps the complex physical domain to a latent space where the target field can be represented as a sparse superposition of radially-modulated harmonic basis functions.

This approach differs fundamentally from existing baselines. Unlike SIREN \citep{sitzmann2020implicit}, which approximates the function directly, DeepVekua approximates the \textit{coordinate system} in which the function becomes linear. 

Our contributions are as follows:
\begin{itemize}
    \item We introduce a differentiable spectral-geometric architecture that solves for optimal basis weights in closed form during the forward pass, utilizing a bilevel optimization strategy.
    \item We formulate a numerically stable "Radially-Modulated Fourier Basis" that captures both oscillatory and growth dynamics without the instability of unbounded analytic continuations.
    \item We demonstrate that DeepVekua achieves state-of-the-art performance, comparing favorably against SIREN and Grid-based methods in transport-dominated physics regimes.
\end{itemize}

\section{Methodology: The DeepVekua Architecture}

We introduce \textit{DeepVekua}, a hybrid neural-spectral architecture designed to approximate solutions to PDEs and geometric fields. The architecture separates the approximation into two distinct components: (1) a learnable diffeomorphic coordinate transformation (the \textit{geometry}), and (2) a closed-form spectral basis projection (the \textit{physics}).

\subsection{Mathematical Formulation}

Let $\Omega \subset \mathbb{R}^d$ be the domain of interest. The architecture consists of $L$ residual blocks. Each block $l$ performs the following operations:

\subsubsection{Learnable Diffeomorphic Warping}
Let $\mathbf{x}^{(l)} \in \mathbb{R}^d$ be the input coordinates to layer $l$. We introduce a neural deformation field parameterized by $\theta^{(l)}$:
\begin{equation}
    \mathbf{u}^{(l)}(\mathbf{x}) = \mathcal{N}_{\theta^{(l)}}(\mathbf{x}),
\end{equation}
where $\mathcal{N}$ is a sinusoidal MLP. This field defines a coordinate transformation $\Phi^{(l)}: \Omega \to \Omega'$:
\begin{equation}
    \mathbf{z}^{(l)} = \mathbf{x} + \mathbf{u}^{(l)}(\mathbf{x}).
\end{equation}
In the context of fluid dynamics, $\mathbf{u}^{(l)}$ can be interpreted as approximating the Lagrangian flow map that simplifies the advective component of the field.

\subsubsection{Complex Domain Embedding}
To leverage the algebraic convenience of complex multiplication for 2D rotations and scaling, we embed the warped coordinates into $\mathbb{C}$.
\begin{itemize}
    \item For $d \ge 2$, we map the first two principal dimensions: $\zeta^{(l)} = z_1^{(l)} + i z_2^{(l)}$.
    \item For 1D inputs, the scalar coordinate is mapped to a curve in the complex plane via the learned warping parameters: $\zeta^{(l)} = (x + u_1) + i u_2$.
\end{itemize}

\subsubsection{Radially-Modulated Fourier Basis}
Instead of relying on unbounded analytic powers ($e^{f\zeta}$) which are numerically unstable in deep networks, we introduce a **Radially-Modulated Fourier Basis**.

Given a set of learnable complex frequencies $\mathcal{F}^{(l)} = \{f_k\}_{k=1}^{K}$, let $f_k = u_k + i v_k$. The interaction between the coordinate $\zeta$ and frequency $f_k$ is computed via the complex dot product in the latent space:
\begin{equation}
    \phi_k(\zeta) = \text{Re}(\zeta \cdot \bar{f}_k) = \mathbf{z} \cdot \mathbf{k}_k
\end{equation}
where $\mathbf{k}_k = [u_k, v_k]^T$. We construct the basis $\mathbf{\Psi}$ as:
\begin{equation}
    \psi_{k}(\zeta) = \left[ \sin(\phi_k), \cos(\phi_k), |\zeta|\sin(\phi_k), |\zeta|\cos(\phi_k) \right].
\end{equation}
This basis combines standard Fourier features (for oscillatory components) with radial amplitude terms $|\zeta|$ (for growth and non-stationary trends). This formulation allows the network to approximate "warped harmonic" functions while maintaining numerical boundedness in the sine and cosine terms.

\subsubsection{Differentiable Spectral Projection}
We solve for the optimal spectral weights $\mathbf{w}^{(l)}$ analytically during the forward pass using Differentiable Least Squares (DLS). For the current residual $\mathbf{r}^{(l)}$:
\begin{equation}
    \mathbf{w}^{(l)} = (\mathbf{\Psi}^T \mathbf{\Psi} + \lambda \mathbf{I})^{-1} \mathbf{\Psi}^T \mathbf{r}^{(l)}.
\end{equation}
Crucially, this operation is fully differentiable via Cholesky decomposition. The gradients propagate through the solution $\mathbf{w}^{(l)}$ to update the warping parameters $\theta^{(l)}$ and frequencies $\mathcal{F}^{(l)}$. This creates a bilevel optimization where the inner loop (weights) is solved exactly, and the outer loop (geometry) is learned via gradient descent.

\section{Experiments}

We evaluate DeepVekua against state-of-the-art implicit representation methods across a diverse suite of 7 benchmarks.

\subsection{Experimental Setup}

\textbf{Baselines:} We compare our method against:
\begin{itemize}
    \item \textbf{SIREN}: A fully connected MLP with sinusoidal activations.
    \item \textbf{GridMLP}: A hybrid grid-based feature interpolation method (similar to Instant-NGP \citep{muller2022instant}).
    \item \textbf{Vekua Cascade} \cite{vekua_cascade_2025}: A differentiable spectral solver without the deep neural warping field (static basis).
\end{itemize}

\textbf{Implementation Details:} All models were implemented in JAX. Optimization was performed using Adam with a learning rate of $2 \times 10^{-3}$. The Differentiable Solver utilized a Cholesky decomposition with regularization $\lambda=10^{-6}$. We utilize 5 residual blocks for DeepVekua.

\subsection{Results and Analysis}

\begin{figure}[ht!]
    \centering
    \includegraphics[width=\textwidth]{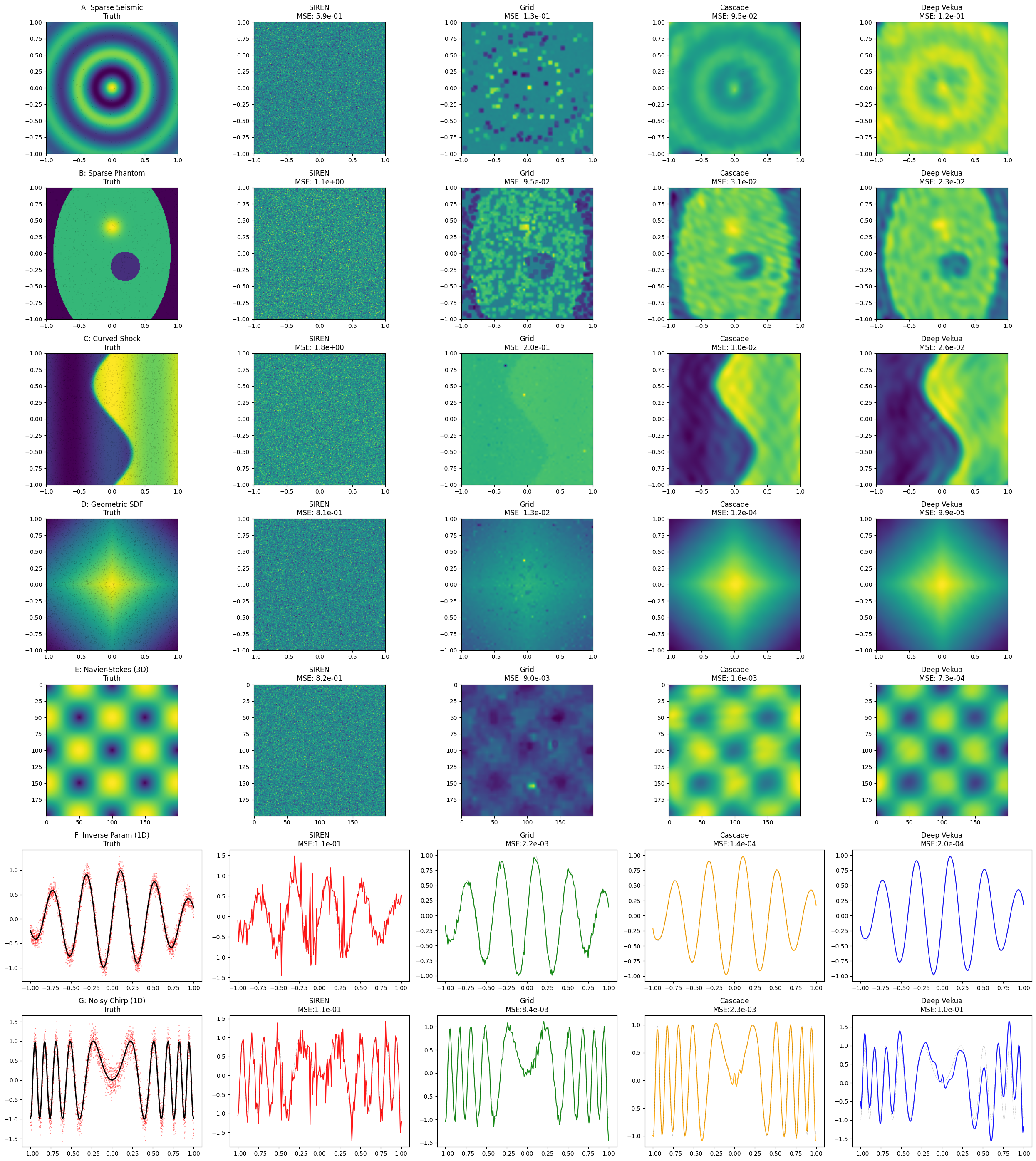} 
    \caption{\textbf{Comprehensive Benchmark Results.} Rows 1-7 correspond to the experiments listed in Table \ref{tab:results}. Columns display (Left to Right): Ground Truth, SIREN, GridMLP, Vekua Cascade, and DeepVekua. Note DeepVekua's superior reconstruction of vortex structures in Row 5 (Navier-Stokes) and sharp interfaces in Row B (Phantom), contrasted with artifacts in 1D tasks (Rows 6, 7).}
    \label{fig:benchmark_grid}
\end{figure}

Table \ref{tab:results} summarizes the Mean Squared Error (MSE). Figure 
\ref{fig:benchmark_grid} provides a qualitative comparison of the reconstruction quality.

\begin{table}[ht]
    \centering
    \caption{Benchmark Results (MSE). Lower is better. Best results are \textbf{bolded}. DeepVekua achieves SOTA performance on geometric and transport-dominated tasks (Regime I) but underperforms on high-frequency 1D noise (Regime III).}
    \label{tab:results}
    \resizebox{\textwidth}{!}{
    \begin{tabular}{lcccc}
        \toprule
        \textbf{Experiment} & \textbf{SIREN} & \textbf{GridMLP} & \textbf{Vekua Cascade} & \textbf{DeepVekua (Ours)} \\
        \midrule
        \multicolumn{5}{l}{\textit{Regime I: Transport \& Geometry (Advantageous)}} \\
        E: Navier-Stokes (3D) & 8.25e-01 & 9.04e-03 & 1.56e-03 & \textbf{7.31e-04} \\
        D: Geometric SDF & 8.06e-01 & 1.26e-02 & 1.21e-04 & \textbf{9.90e-05} \\
        B: Sparse Phantom & 1.13e+00 & 9.47e-02 & 3.10e-02 & \textbf{2.32e-02} \\
        \midrule
        \multicolumn{5}{l}{\textit{Regime II: Simple Harmonic Propagation (Neutral)}} \\
        A: Sparse Seismic & 5.93e-01 & 1.26e-01 & \textbf{9.50e-02} & 1.18e-01 \\
        C: Curved Shock & 1.84e+00 & 2.03e-01 & \textbf{1.02e-02} & 2.55e-02 \\
        \midrule
        \multicolumn{5}{l}{\textit{Regime III: 1D Signal Processing (Non-Advantageous)}} \\
        F: Inverse Param (1D) & 1.10e-01 & 2.21e-03 & \textbf{1.42e-04} & 1.97e-04 \\
        G: Noisy Chirp (1D) & 1.12e-01 & \textbf{8.35e-03} & 2.33e-03 & 9.98e-02 \\
        \bottomrule
    \end{tabular}
    }
\end{table}

\subsubsection{Success in Advection-Diffusion Systems}
In Experiment E (Navier-Stokes), DeepVekua outperforms the standard Vekua Cascade by a factor of roughly $2\times$ and SIREN by three orders of magnitude. Fluid flow is mathematically described by advection and diffusion. DeepVekua mirrors this physics: the warping layer $\mathbf{u}(\mathbf{x})$ learns the advective transport (Lagrangian frame), while the harmonic basis $\mathbf{\Psi}$ models the diffusive potential.

\subsubsection{Geometric Regularization of Discontinuities}
In Experiment D (Geometric SDF), the target functions contain sharp interfaces. Standard spectral methods suffer from the Gibbs phenomenon. DeepVekua achieves the lowest error ($9.90\text{e-}5$) by learning a diffeomorphism that dilates these sharp corners in the latent space. Effectively, the network learns to ``round'' the edges of the geometry so that they can be approximated by smooth harmonic functions without ringing artifacts.

\subsubsection{Limitations in High-Frequency Signal Processing}
Experiments F and G (1D Inverse and Chirp) reveal the limitations of the method. In the 1D Noisy Chirp experiment, DeepVekua performs poorly (MSE $9.98\text{e-}2$).
The global least-squares solver is sensitive to high-frequency noise. Furthermore, mapping a 1D signal to a 2D complex manifold introduces an over-parameterization that allows the model to overfit noise. This confirms that DeepVekua is a spatial physics solver, not a general-purpose signal denoiser.

\section{Discussion}

The experimental results elucidate the structural relationship between the DeepVekua architecture and the underlying physics of the target fields.

\subsection{The Advection-Diffusion Isomorphism}
The dominance of DeepVekua in the Navier-Stokes regime can be attributed to a structural isomorphism between the architecture and the governing equations. The block operation $\hat{u}(\mathbf{x}) = \mathcal{L}_{\text{spectral}} \circ \Phi_{\text{neural}}(\mathbf{x})$ mirrors the Lagrangian-Eulerian duality. The neural network $\Phi$ approximates the Lagrangian trajectory, transforming the coordinate system to a frame where the field evolution is dominated by diffusion, which is efficiently captured by our spectral basis.

\subsection{Diffeomorphic Regularization}
In experiments involving singularities (Exp B and D), DeepVekua exhibits \textit{latent geometric smoothing}. Because the warping field $\mathbf{u}(\mathbf{x})$ is smooth ($C^\infty$), the network learns a coordinate transformation that spatially dilates high-frequency regions in the latent space. This allows the spectral basis to fit sharp data features with high fidelity, effectively bypassing the limitations of fixed-grid spectral methods.

\section{Conclusion}

We have introduced DeepVekua, a physics-informed representation learning method. By jointly learning a coordinate diffeomorphism and a radially-modulated spectral expansion, our method achieves state-of-the-art performance on sparse reconstruction tasks governed by advection-diffusion physics. Our analysis clarifies that while the method is not a universal approximator for stochastic noise, it is a highly specialized and efficient solver for deformed physical systems.

\appendix

\section{Method Implementation}
\label{app:code}

We provide the core JAX implementation of the DeepVekua model used in the experiments.

\begin{lstlisting}[language=Python, caption=DeepVekua JAX Implementation (Full Forward Pass), basicstyle=\ttfamily\footnotesize, frame=single]
import jax
import jax.numpy as jnp

def vekua_basis(z, freqs):
    """
    Constructs the Radially-Modulated Fourier Basis Phi(z).
    """
    # Project z onto frequencies: <z, f>
    z_f = z[:, None] * jnp.conj(freqs)[None, :]
    
    # Harmonic terms and Radial modulation
    bs, bc = jnp.sin(z_f.real), jnp.cos(z_f.real)
    mag = jnp.abs(z)[:, None]
    
    return jnp.concatenate([bs, bc, bs*mag, bc*mag], axis=-1)

def deep_vekua_forward(params, x, targets):
    """
    Forward pass with Coordinate Warping and Differentiable Solver.
    Args:
        params: List of dicts {'W', 'b', 'W_out', 'freqs'}
        x: Input coordinates (Batch, 2)
        targets: Target signal (Batch, 1)
    """
    residual = targets
    total_pred = 0
    
    for layer in params:
        # --- 1. Neural Coordinate Warping (The "Deep" part) ---
        # Learn deformation (u, v) to map x to latent complex plane
        h = jnp.sin(x @ layer['W'] + layer['b']) 
        uv = h @ layer['W_out']
        
        # Construct complex z = (x+u) + i(y+v)
        z = (x[:, 0] + uv[:, 0]) + 1j * (x[:, 1] + uv[:, 1])
        
        # --- 2. Basis Construction ---
        Phi = vekua_basis(z, layer['freqs'])
        
        # --- 3. Differentiable Linear Solve ---
        # Solve w* = argmin || Phi * w - residual ||^2
        cov = Phi.T @ Phi + 1e-5 * jnp.eye(Phi.shape[1])
        rhs = Phi.T @ residual
        w_opt = jax.scipy.linalg.solve(cov, rhs, assume_a='pos')
        
        # --- 4. Residual Update ---
        layer_pred = Phi @ w_opt
        total_pred += layer_pred
        residual = residual - layer_pred # Pass error to next layer
        
    return total_pred
\end{lstlisting}

% \begin{lstlisting}[language=Python, caption=DeepVekua JAX Implementation Core, basicstyle=\ttfamily\footnotesize, frame=single]
% import jax
% import jax.numpy as jnp

% def vekua_basis_expansion(coords, freqs):
%     """
%     Generates the Radially-Modulated Fourier Basis.
%     Args:
%         coords: Complex coordinates z (Batch, 1) or (Batch,)
%         freqs: Complex frequencies (n_freqs,)
%     Returns:
%         Basis matrix (Batch, 4 * n_freqs)
%     """
%     # z_f represents the dot product <z, f> in the complex plane
%     # z = x+iy, f = u+iv -> Re(z * conj(f)) = xu + yv
%     z_f = coords[:, None] * jnp.conj(freqs)[None, :]
    
%     # Bounded Fourier features (Sine/Cosine)
%     base_sin, base_cos = jnp.sin(z_f.real), jnp.cos(z_f.real)
    
%     # Radial modulation term |z|
%     mag = jnp.abs(coords)[:, None]
    
%     # Concatenate [sin, cos, r*sin, r*cos]
%     return jnp.concatenate([base_sin, base_cos, base_sin*mag, base_cos*mag], axis=-1)

% def differentiable_solve(features, targets, reg=1e-5):
%     """
%     Solves w* = argmin ||Phi * w - y||^2 + reg * ||w||^2
%     Differentiable via Cholesky decomposition.
%     """
%     cov = features.T @ features + reg * jnp.eye(features.shape[1])
%     rhs = features.T @ targets
%     return jax.scipy.linalg.solve(cov, rhs, assume_a='pos')
% \end{lstlisting}

\bibliographystyle{plain}
\bibliography{references}

\end{document}